\documentclass[
]{ceurart}

  \makeatletter                                                                                                             
  \def\@copyrightLine{}                      
  \RenewDocumentCommand \nonumnote { m } {}
                                  
  \makeatother

\usepackage{listings}

\usepackage{siunitx}
\usepackage{booktabs}
\usepackage{graphicx}
\usepackage{pifont}
\usepackage{multirow}
\usepackage{xcolor}
\usepackage{colortbl}
\usepackage{tikz}
\usetikzlibrary{arrows.meta,positioning,calc}
\begin{document}

\copyrightyear{2026}
\copyrightclause{Copyright for this paper by its authors.
  Use permitted under Creative Commons License Attribution 4.0
  International (CC BY 4.0).}

% \conference{Woodstock'22: Symposium on the irreproducible science,
%   June 07--11, 2022, Woodstock, NY}

% \conference{LIR'22: 1st Late Interaction Workshop (LIR) @ ECIR 2026,
%   March 29--April 02, 2026, Delft, The Netherlands}
\title{ColBERT-Zero: To Pre-train Or Not To Pre-train ColBERT models}

% \tnotetext[1]{You can use this document as the template for preparing your
%   publication. We recommend using the latest version of the ceurart style.}

\author[1]{Antoine Chaffin}[%
orcid=0000-0003-3605-4097,
email=antoine.chaffin@lighton.ai,
url=https://antoine.chaffin.fr/,
]
\cormark[1]
\fnmark[1]
\address[1]{LightOn, France}
% \address[2]{Joint Institute for Nuclear Research,
%   6 Joliot-Curie, Dubna, Moscow region, 141980, Russian Federation}

\author[2]{Luca Arnaboldi}[%
orcid=0009-0001-9739-8849,
email=luca.arnaboldi@epfl.ch,
url=https://www.arnaboldi.lu,
]
\cormark[1]
\fnmark[1]
% \address[3]{Vrije Universiteit Amsterdam, De Boelelaan 1105, 1081 HV Amsterdam, The Netherlands}
\address[2]{Ecole Polytechnique Fédérale de Lausanne (EPFL), IdePHICS Lab, CH-1015 Lausanne, Switzerland}

\author[1]{Amélie Chatelain}[%
orcid=0000-0002-7562-4913,
email=amelie@lighton.ai,
url=https://meet.ameliechatelain.com/,
]

\author[2]{Florent Krzakala}[%
orcid=0000-0002-9421-8566,
email=florent.krzakala@epfl.ch,
url=https://florentkrzakala.com,
]

\cortext[1]{Corresponding author.}
\fntext[1]{These authors contributed equally.}

\begin{abstract}
  Current state-of-the-art multi-vector models are obtained through a small Knowledge Distillation (KD) training step on top of strong single-vector models, leveraging the large-scale pre-training of these models. In this paper, we study the pre-training of multi-vector models and show that large-scale multi-vector pre-training yields much stronger multi-vector models. Notably, a fully ColBERT-pre-trained model, ColBERT-Zero, trained only on public data, outperforms GTE-ModernColBERT as well as its base model, GTE-ModernBERT, which leverages closed and much stronger data, setting new state-of-the-art for model this size. We also find that, although performing only a small KD step is not enough to achieve results close to full pre-training, adding a supervised step beforehand allows to achieve much closer performance while skipping the most costly unsupervised phase. Finally, we find that aligning the fine-tuning and pre-training setups is crucial when repurposing existing models. To enable exploration of our results, we release \href{https://huggingface.co/collections/lightonai/colbert-zero}{various checkpoints} as well as \href{https://github.com/lightonai/pylate/tree/main/examples/train/ColBERT-zero/}{code used to train them}.
\end{abstract}

\begin{keywords}
  ColBERT \sep
  Pre-training
\end{keywords}

\maketitle

\section{Introduction}
Late interaction models, also referred to as multi-vector or ColBERT~\cite{DBLP:conf/sigir/KhattabZ20} models, have gained popularity thanks to their out-of-domain~\cite{DBLP:conf/naacl/SanthanamKSPZ22}, long-context~\cite{GTE-ModernColBERT}, and reason-intensive~\cite{Reason-ModernColBERT} retrieval capabilities. Despite these clear advantages, they are less studied than their single-vector counterparts. Thus, the state-of-the-art ColBERT models such as GTE-ModernColBERT~\cite{GTE-ModernColBERT} and ColBERT-small~\cite{ColBERT-small} are built by performing a small Knowledge Distillation (KD) phase on top of a strong pre-trained single-vector (dense) model. 
Even mxbai-edge-colbert-v0~\cite{DBLP:journals/corr/abs-2510-14880}, a recent study that details a recipe to train very strong ColBERT models from scratch (that is, MLM-pretained Ettin~\cite{weller2025seqvsseqopen} models) performed all the early stages in a single-vector setting and only used multi-vector in the latest stage, knowledge distillation.

\begin{figure}[t]
  \centering
  \resizebox{\textwidth}{!}{\input{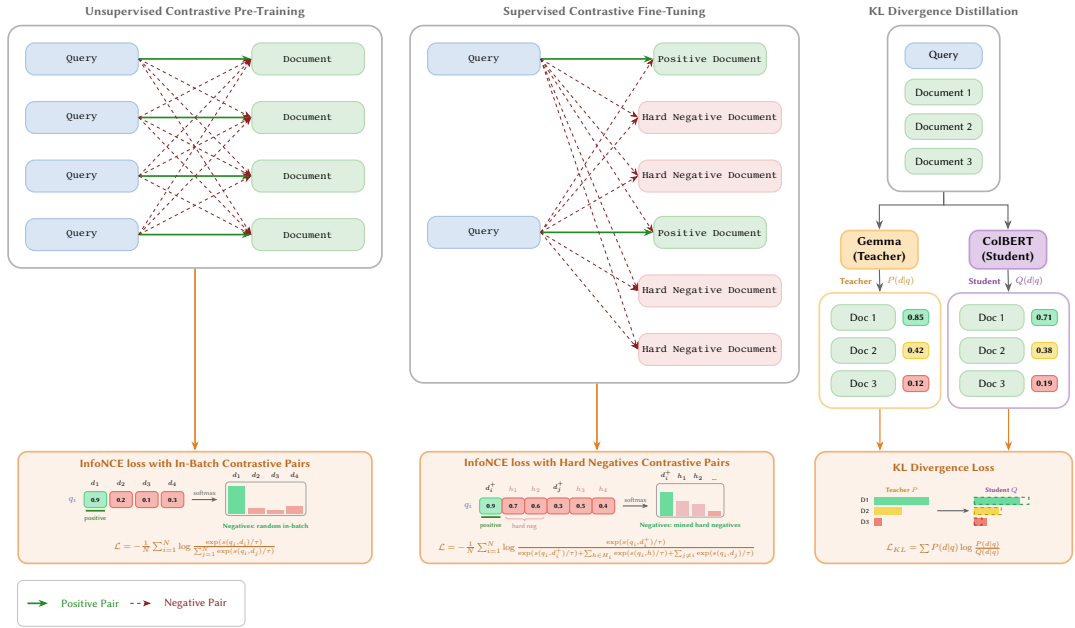}}
  \caption{An illustration of the three training phases. Unsupervised contrastive pre-training is a very large scale training relying exclusively on large batch sizes to use in-batch negatives. Supervised contrastive fine-tuning is a refinement step that leverages mined hard negative to provide a stronger signal. The knowledge distillation step use a strong teacher to scores various documents and use KL divergence to make the student distribution fits the teachers'. }
  \label{fig:training_phase}
\end{figure}

Training a retrieval model typically involves up to three phases (illustrated in Figure~\ref{fig:training_phase}): (1) \emph{unsupervised contrastive pre-training}, a large-scale contrastive phase relying on in-batch negatives; (2) \emph{supervised contrastive fine-tuning}, which refines the model using mined hard negatives; and (3) \emph{Knowledge Distillation} (KD), where a strong teacher's relevance scores guide the student via KL divergence. 
Although applying only KD yields strong models and allows us to leverage various strong dense models, it treats the multi-vector setting as an afterthought. This approach could be suboptimal compared to performing contrastive steps in the multi-vector setting before the knowledge distillation step. 

In this paper, we investigate which training steps are necessary for optimal ColBERT performance. Specifically, we ask two questions: is knowledge distillation alone sufficient to transfer dense model quality to the multi-vector setting? And if not, can a supervised contrastive phase before KD close the gap without the costly unsupervised pre-training? To answer these, we compare models trained with (a) the knowledge distillation step alone, (b) the supervised contrastive and knowledge distillation steps, and (c) all steps, including the large-scale unsupervised contrastive step; see Figure~\ref{fig:pipelines} for details. Our findings show that KD alone is insufficient for optimal performance, but that adding a supervised phase beforehand yields competitive results without the most expensive unsupervised phase, particularly when aligning the fine-tuning setup with the pre-training one. Indeed, we find that aligning the use of prompts is crucial when repurposing existing models, with misalignment leading to significant performance degradation The resulting model, \emph{ColBERT-Zero}, trained only on public data, outperforms both the state-of-the-art GTE-ModernColBERT as well as its base model, GTE-ModernBERT, trained on closed and much stronger data, and sets new state-of-the-art for its size. To support reproducibility and further exploration, we release \href{https://huggingface.co/collections/lightonai/colbert-zero}{all models}, including intermediate checkpoints with various training parameters, as well as the \href{https://github.com/lightonai/pylate/tree/main/examples/train/ColBERT-zero/}{full training scripts}.

\section{Scaling Pre-Training of ColBERT Models to Nomic Embed Scale}
In order to establish whether pre-training ColBERT models is beneficial over a simple distillation, we propose to pre-train a ColBERT model on the widely-known datasets used by Nomic Embed~\cite{DBLP:journals/tmlr/NussbaumMMD25}. 
This approach has several advantages: first, it is a robust choice of public data that is easily accessible through the Hugging Face dataset library, which also facilitates implementation. Secondly, a ModernBERT dense model has been trained on this mixture, allowing for easy comparison.

\begin{figure}[t]
  \centering
  \resizebox{\textwidth}{!}{\definecolor{mlmcolor}{HTML}{4A4A4A}
\definecolor{densecolor}{HTML}{5B9BD5}
\definecolor{colbertcolor}{HTML}{ED7D31}
\definecolor{arrowgray}{HTML}{666666}
\begin{tikzpicture}[
    node distance=1.8cm and 2.2cm,
    box/.style={
        rectangle,
        rounded corners=3pt,
        minimum width=2.4cm,
        minimum height=0.9cm,
        align=center,
        font=\small,
        draw=#1,
        line width=0.8pt,
        fill=#1!12
    },
    dense/.style={box=densecolor},
    colbert/.style={box=colbertcolor},
    mlm/.style={box=mlmcolor},
    arr/.style={-{Stealth[length=2.5mm]}, line width=0.7pt, arrowgray},
    pipelabel/.style={font=\small\bfseries, anchor=west},
    costlabel/.style={font=\small\itshape, anchor=west, text=arrowgray},
]
% === Starting point ===
\node[mlm] (mlm) {MLM\\Pre-training};
% === Pipeline A: KD only (top) ===
\node[dense, right=of mlm, yshift=2.2cm] (a1) {Unsupervised};
\node[dense, right=of a1] (a2) {Supervised};
\node[colbert, right=of a2] (a3) {KD};
% === Pipeline B: Sup + KD (middle) ===
\node[dense, right=of mlm] (b1) {Unsupervised};
\node[colbert, right=of b1] (b2) {Supervised};
\node[colbert, right=of b2] (b3) {KD};
% === Pipeline C: Full ColBERT (bottom) ===
\node[colbert, right=of mlm, yshift=-2.2cm] (c1) {Unsupervised};
\node[colbert, right=of c1] (c2) {Supervised};
\node[colbert, right=of c2] (c3) {KD};
% === Arrows from MLM ===
\draw[arr] (mlm.east) -- ++(0.4,0) |- (a1.west);
\draw[arr] (mlm.east) -- (b1.west);
\draw[arr] (mlm.east) -- ++(0.4,0) |- (c1.west);
% === Pipeline A arrows ===
\draw[arr] (a1) -- (a2);
\draw[arr] (a2) -- (a3);
% === Pipeline B arrows ===
\draw[arr] (b1) -- (b2);
\draw[arr] (b2) -- (b3);
% === Pipeline C arrows ===
\draw[arr] (c1) -- (c2);
\draw[arr] (c2) -- (c3);
% === Pipeline labels + compute costs (right side) ===
\node[pipelabel] at ($(a3.east)+(0.3,0.12)$) {(a) KD only};
\node[costlabel] at ($(a3.east)+(0.3,-0.22)$) {{\raise.17ex\hbox{$\scriptstyle\sim$}}8 GH200-hours};
\node[pipelabel] at ($(b3.east)+(0.3,0.12)$) {(b) Supervised + KD};
\node[costlabel] at ($(b3.east)+(0.3,-0.22)$) {{\raise.17ex\hbox{$\scriptstyle\sim$}}40 GH200-hours};
\node[pipelabel] at ($(c3.east)+(0.3,0.12)$) {(c) ColBERT-Zero};
\node[costlabel] at ($(c3.east)+(0.3,-0.22)$) {{\raise.17ex\hbox{$\scriptstyle\sim$}}408 GH200-hours};
% === Legend ===
\node[dense, minimum width=1.4cm, minimum height=0.55cm] (leg1) at ($(c1.south)!0.5!(c2.south)+(0,-1.0)$) {};
\node[anchor=west, font=\small] at (leg1.east) {~Dense};
\node[colbert, minimum width=1.4cm, minimum height=0.55cm] (leg2) at ($(leg1.east)+(2.4,0)$) {};
\node[anchor=west, font=\small] at (leg2.east) {~ColBERT};
\end{tikzpicture}}
  \caption{The different training pipelines compared in this work. KD only is the most common setup, while ColBERT-Zero is the most expensive. Supervised + KD is an in-between much cheaper than the full pre-training.}
  \label{fig:pipelines}
\end{figure}
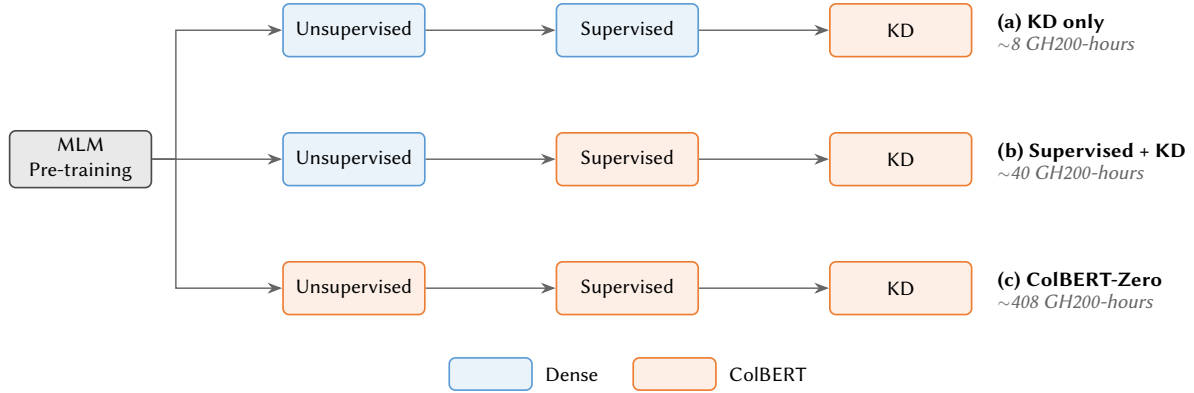

We thus train various models to explore the effect of each training phase: unsupervised contrastive, supervised contrastive, and knowledge distillation. All the different settings are illustrated in Figure~\ref{fig:pipelines}.
The baseline is a simple knowledge distillation step performed on top of the ModernBERT-embed model that has undergone both unsupervised and supervised contrastive phases in the dense setting and thus only a KD phase in the ColBERT setting (setting (a)). It is the most usual ColBERT training setup.
This model is compared to ColBERT-Zero, for which both the unsupervised and supervised steps are also performed in the multi-vector setting, i.e, the whole pre-training is run in the multi-vector setting (setting (c)). This allows for a fair comparison as both models use the same original base model and the exact same data, the only difference being whether the contrastive phases are performed in the dense or late interaction setting.
As the authors also released the dense model obtained after the unsupervised phase only, we study additionally the impact of skipping the unsupervised ColBERT pre-training but still performing the supervised contrastive phase in the ColBERT setting before KD, allowing to study whether the large-scale unsupervised phase is required or simply adding the supervised phase is enough (setting (b)). As the unsupervised phase is, by far, the most expensive step (almost ten times more than supervised + KD, see Table~\ref{tab:hyperparams}), being able to leverage the one already performed by the dense model would be very interesting. 

%An illustration of the different training phases can be found in Figure~\ref{fig:training_phase} and one of the three different training pipeline compared can be found in Figure~\ref{fig:pipelines}.

We use standard setups for each phase, training for only one epoch each time.
For the unsupervised pre-training, we use the large dataset from Nomic Embed and rely on in-batch negatives. The InfoNCE loss temperature was designated as a learnable parameter with the objective of ascertaining the optimal value for the dataset, and later fixed during training. 
For the supervised phase, we leverage smaller and higher quality datasets with negatives mined by Nomic, while the optimal temperature was determined similarly to the unsupervised phase.
The knowledge distillation setup follows the one of GTE-ModernColBERT, using the MS-MARCO dataset with mined samples scored by \href{https://huggingface.co/BAAI/bge-reranker-v2-gemma}{bge-reranker-v2-gemma}. All the hyperparameters are summarized in Table~\ref{tab:hyperparams}, including the learning rate sweep ranges used for each phase. We use the NanoBEIR evaluator to choose the best values for temperature and learning rate as running the full BEIR~\cite{DBLP:journals/corr/abs-2104-08663} is expensive.
%The query and document lenghts are the same as the supervised phase. Unsupervised pre-training is done on 8 nodes of 4 GH200, while supervised fine-tuning and knowledge distillation use only one node.
\begin{table*}
\centering
\caption{
  Training hyperparameters of the 3 different phases of learning. \\ 
  The learning rates are swept logarithmically in the given ranges, testing 10 values for each phase.
  The query and document lengths include the tokens of the prompts (see Section~\ref{sec:prompts}).
  GPU hours are measured on a cluster of GH200 GPUs, where nodes have 4 GPU each; unsupervised is done on 8 nodes, while supervised and KD on 1 node.
}
\label{tab:hyperparams}
\begin{tabular}{l c c c c c c}
\toprule
\textbf{Phase} & \textbf{LR sweep range} & \textbf{Batch size} & \textbf{Temp.} & \textbf{Query length} & \textbf{Doc. length} & \textbf{GPU hours}\\
\midrule
% Unsupervised            & \(\num{3e-3}, \num{1e-3}, \num{8e-4}, \num{6e-4}, \num{4e-4}, \num{3e-4}, \num{2e-4}, \num{1e-4}, \num{3e-5}, \num{1e-5}\) & 16384 & 0.2\\
% Supervised              & \(\num{2e-5}, \num{1e-5}, \num{2e-6}, \num{1e-6}, \num{8e-7}, \num{6e-7}, \num{4e-7}, \num{2e-7}, \num{1e-7}, \num{8e-8}\) & 64 & 0.2 \\
% KD  & \(\num{1e-3}, \num{8e-4}, \num{3e-4}, \num{1e-4}, \num{6e-5}, \num{4e-5}, \num{2e-5}, \num{1e-5}, \num{3e-7}, \num{1e-7}\) & \(128^\star\) & n.a.\\
Unsupervised  & \([\num{3e-3}, \num{1e-5}]\) & 16384 & 0.2 & 39 & 187 & 368\\
Supervised    & \([\num{2e-5}, \num{8e-8}]\) & 64 & 0.2 & 39 & 519 & 32\\
KD            & \([\num{1e-3},  \num{1e-7}]\) & \(128^\star\) & n.a. & 39 & 519 & 8\\
\bottomrule
\end{tabular}
\vspace{.1cm}

{\small \(\star\) Effective batch size achieved with gradient accumulation of 2.}
\end{table*}

We use the PyLate~\cite{DBLP:conf/cikm/ChaffinS25} library and leverage the implementation of GradCache~\cite{DBLP:conf/rep4nlp/GaoZHC21} to scale the per-GPU batch size arbitrarily without VRAM constraints (as standard gradient accumulation is not applicable for contrastive learning). We also gather the samples from the other GPUs in a multi-GPU setting to leverage all computed representations and scale batch size without additional memory requirement. These two features combined allows to easily reach batch sizes of 16k, which is required for unsupervised training to get plausible in-batch hard negatives~\cite{10.5555/3524938.3525087}. Finally, each batch is composed of only one data source at a time, to prevent shortcut learning~\cite{DBLP:journals/tmlr/NussbaumMMD25}. We release the checkpoints obtained after each phase (unsupervised, supervised, distilled) for all pipelines, enabling the community to use them as starting points for their own experiments.

% Unsupervised pre-training is done on 8 nodes of 4 GH200, while supervised fine-tuning and knowledge distillation use only one node. The total GH200 hours of each phase are respectively, 368, 32 and 8, highlighting that the unsupervised pre-training is --by far-- the most expensive phase. 

% It is worth noting that, besides the limited efforts put into training of ColBERT models (w.r.t dense), one of the reason of the lack of pre-trained ColBERT was the lack of available resources to easily explore these setups.
% Thankfully, PyLate~\cite{DBLP:conf/cikm/ChaffinS25} offers different features that make it suitable for running such large-scale pre-training.
% First, the implementation of GradCache~\cite{DBLP:conf/rep4nlp/GaoZHC21} allows to scale the per-GPU batch size arbitrarily without VRAM constraints (as standard gradient accumulation is not applicable for contrastive learning). Then, the option of gathering the samples from the other GPUs in a multi-GPUs settings allows to leverage all the computed representations to scale batch size without additional memory requirement. These two features combined allows to easily reach batch sizes of 16k, which is required for unsupervised training to get plausible in-batch hard negatives.
% Finally, PyLate allows to easily track in-training evaluations such as NanoBEIR on Weights\&Biases, which makes sweeping and monitoring training easier.

\section{Results}
\subsection{Main Results}
\begin{table*}[t]
\centering
\caption{Retrieval performance (nDCG@10) across BEIR benchmark datasets.}
\label{tab:original-results}
\resizebox{\textwidth}{!}{%
\begin{tabular}{l c ccccccccccccccc}
\toprule
\textbf{Model} & \textbf{Avg} & \rotatebox{90}{\textbf{FiQA}} & \rotatebox{90}{\textbf{NFCorpus}} & \rotatebox{90}{\textbf{TREC-COVID}} & \rotatebox{90}{\textbf{Touche}} & \rotatebox{90}{\textbf{ArguAna}} & \rotatebox{90}{\textbf{Quora}} & \rotatebox{90}{\textbf{SCIDOCS}} & \rotatebox{90}{\textbf{SciFact}} & \rotatebox{90}{\textbf{NQ}} & \rotatebox{90}{\textbf{ClimateFEVER}} & \rotatebox{90}{\textbf{HotpotQA}} & \rotatebox{90}{\textbf{DBPedia}} & \rotatebox{90}{\textbf{CQADupstack}} & \rotatebox{90}{\textbf{FEVER}} & \rotatebox{90}{\textbf{MSMARCO}} \\
\midrule
\multicolumn{17}{l}{\textit{Baselines}} \\
\midrule
ModernBERT-embed-unsupervised & 47.05 & 42.53 & 35.33 & 68.44 & 18.58 & 48.82 & 88.63 & 19.83 & 72.30 & 46.32 & 22.97 & 60.00 & 37.97 & 42.40 & 67.39 & 34.23 \\
ModernBERT-embed-supervised & 52.89 & 40.59 & 33.40 & \textbf{84.15} & 31.91 & 48.96 & \textbf{88.85} & 18.59 & 69.63 & 62.15 & 35.67 & 67.11 & 41.50 & 42.08 & 87.35 & 41.47 \\
GTE-ModernColBERT & 54.67 & 45.28 & \textbf{37.93} & 83.59 & 31.23 & 48.51 & 86.61 & 19.06 & 76.34 & 61.80 & 30.62 & 77.32 & 48.03 & 41.00 & 87.44 & 45.32 \\
gte-modernbert & 55.33 & \textbf{48.81} & 36.44 & 81.95 & 21.68 & \textbf{72.68} & 88.55 & 21.29 & \textbf{77.40} & 57.62 & \textbf{37.74} & 69.47 & 41.79 & 42.63 & \textbf{91.03} & 40.90 \\
\midrule
\multicolumn{17}{l}{\textit{KD from dense supervised}} \\
\midrule
embed-base distilled & 54.09 & 42.51 & 37.01 & 79.52 & 34.58 & 51.75 & 87.67 & 18.15 & 75.04 & 61.45 & 28.31 & 76.70 & 47.54 & 40.68 & 84.82 & 45.57 \\
\midrule
\multicolumn{17}{l}{\textit{Supervised + KD from dense unsupervised}} \\
\midrule
embed-base supervised & 50.72 & 40.09 & 35.56 & 71.12 & 25.53 & 44.27 & 86.96 & 18.19 & 73.78 & 58.89 & 32.95 & 71.49 & 43.23 & 42.55 & 70.51 & 45.72 \\
embed-base distilled & 55.12 & 41.50 & 36.51 & 77.46 & \textbf{33.77} & 52.45 & 86.26 & 18.66 & 74.90 & \textbf{62.24} & 37.27 & \textbf{80.07} & \textbf{48.27} & 41.60 & 89.71 & \textbf{46.17} \\
\midrule
\multicolumn{17}{l}{\textit{ColBERT-Zero} (full pre-training)} \\
\midrule
Unsupervised & 51.44 & 45.38 & 36.88 & 67.82 & 22.59 & 51.53 & 87.78 & \textbf{22.30} & 76.76 & 58.80 & 24.24 & 68.29 & 43.16 & \textbf{45.76} & 81.58 & 38.78 \\
Supervised & 51.81 & 42.45 & 35.60 & 74.72 & 23.83 & 41.81 & 87.19 & 19.85 & 73.71 & 61.95 & 35.01 & 71.37 & 46.20 & 45.16 & 72.61 & 45.68 \\
Distilled & \textbf{55.43} & 42.62 & 37.28 & 78.69 & 36.13 & 53.07 & 85.24 & 19.88 & 76.50 & 61.66 & 35.72 & 79.41 & 47.48 & 41.34 & 90.59 & 45.80 \\
\bottomrule
\end{tabular}%
}
\end{table*}

We evaluate the different models on the BEIR~\cite{DBLP:journals/corr/abs-2104-08663} benchmark and report the nDCG@10 values in Table~\ref{tab:original-results}. We also report several models sharing the same backbone (ModernBERT-base) to provide comparison. Specifically, we report the dense model from Nomic (ModernBERT-embed), the state-of-the-art late interaction model GTE-ModernColBERT as its base model, gte-modernbert. 

Our first finding is that, the fully pre-trained ColBERT model outperforms not only the state-of-the-art late interaction model GTE-ModernColBERT but also its base model, gte-modernbert. This is very strong result, as those are performant models optimized on strong non-public data. The superiority of the GTE data over the Nomic one is shown by the huge gap in BEIR results for the dense models (52.9 for Nomic data vs 55.33 for GTE data). The significance of this result is best appreciated in context: at the time of writing\footnote{January 2026} the model is ranked as the top-1 model on the \href{https://huggingface.co/spaces/mteb/leaderboard}{MTEB BEIR leaderboard}~\cite{muennighoff2022mteb} for model <150M parameters (top-4 for models <500M). Consequently,  this highlights that \textbf{running a full ColBERT pre-training before performing knowledge distillation is noticeably better than only performing KD on top of a pre-trained dense model}. This is further confirmed by the results obtained by performing the KD step on top of the dense supervised model from Nomic, lagging much behind the fully ColBERT-pre-trained model by more than 1.3 pts of nDCG@10 despite leveraging the same data.

Having established that KD alone is insufficient, we now explore whether a supervised contrastive phase before KD can close this gap without the costly unsupervised pre-training. Starting from the unsupervised dense model and running both the supervised and knowledge distillation phases in the ColBERT setting, we observe a gain of more than 1 nDCG@10 over KD alone, nearly matching the fully pre-trained ColBERT (0.31 difference).  This demonstrate that this setup is a \textbf{highly efficient alternative that skips the most computationally intensive training stage while retaining nearly all the performance benefits}. This costs 40 GH200-hours instead of 408: \textbf{roughly 10× cheaper for 99.4\% of the performance}. While the full three-step pipeline retains a slight edge, the difference is limited, particularly considering the potential for noise from BEIR, and is likely to diminish further with heavier fine-tuning, as discussed in Section~\ref{sec:discussions}.

\subsection{Impact of the Prompt}
\label{sec:prompts}
It is important to note that Nomic's dense models are pre-trained using specific prompts ( ``\texttt{search\_query:}" and ``\texttt{search\_document:}"). Unlike frameworks like Promptriever~\cite{DBLP:conf/iclr/WellerDLPZH25}, which use prompts to enable task-specific ``promptability", Nomic utilizes them to enforce an asymmetric encoding. While ColBERT features a native asymmetric mechanism via [Q] and [D] markers, a mismatch between the pre-training prompts and our fine-tuning setup could undermine performance. To ensure a fair comparison, all models reported in the previous section utilize these prompts. However, we experienced with a ``prompt-free" ColBERT-Zero model (i.e, full pre-training) and observed a significant performance drop, suggesting that prompts provide benefits beyond mere identification. To explore this phenomenon, we ran various training with and without prompts at different steps of the training and also explore the impact of increasing the query and document lengths to account for the additional prompts\footnote{During training, as increasing the lengths during evaluation introduce little to no change: we cover the maximum lengths of the BEIR datasets}. As the prompts are both 7 tokens long, we increased the query/documents lengths by 7.

As shown in Table~\ref{tab:prompts-main}, our results highlights that aligning the fine-tuning setup with the pre-training (by using prompts) is crucial; our ablation studies confirm that misalignment leads to significant performance degradation. Incorporating prompts during the fine-tuning of pre-trained models with prompts enhances results, while including them for models pre-trained without prompts leads to a decline. However, even when the pre-training and fine-tuning setups are aligned, the model leveraging more explicit prompts still significantly outperforms the other. This does suggest that those prompts are beneficial. That said, it is hard to properly disentangle all the possible reasons explaining the differences in performance as there are different effect at play. We conjecture this may be a form of implicit \emph{query expansion}, a mechanism that has shown very useful in the early variant of ColBERT~\cite{DBLP:conf/sigir/KhattabZ20}. Typically, these tokens serve as ``placeholder" tokens that does not correspond to a specific semantic token but can be used by the model to store global information about the sequence. Despite being very helpful, this mechanism is not used in modern models leveraging Flash Attention\footnote{Query expansion worked by using PAD tokens that were not attended by other tokens in the sequence. In the original (not optimized) attention implementations, all the attentions scores were computed and the contributions of masked tokens were zeroed out at the end so that the representations of the tokens were only computed w.r.t unmasked tokens. However, the representations of the masked tokens w.r.t all the others tokens were still computed. Those are usually thrown away using attention mask afterwards, but in the case of ColBERT, those were used as query expansion. With more recent and optimized implementations of attention such as Flash Attention, the embedding of masked tokens are zeros/NaNs, preventing their usage as query expansion.}~\cite{dao2022flashattention} such as ModernBERT~\cite{DBLP:conf/acl/WarnerCCWHTGBLA25}, the base model used by all models in our experiments. Our preliminary analysis (detailed in Appendix~\ref{appendix:prompts}) suggests that the performance gains stem from a joint effect of prompts and query/document lengths, pointing to a richer interaction between these factors; to facilitate further investigation of this interaction, we release checkpoints of models trained both with and without prompts at each stage.

\begin{table*}[t]
\centering
\caption{Retrieval performance (nDCG@10) across BEIR benchmark datasets when performing supervised + KD phases on top of unsupervised models pre-trained with or without prompts and the impact of adding the prompts during fine-tuning.}
\label{tab:prompts-main}
\resizebox{\textwidth}{!}{%
\begin{tabular}{l c ccccccccccccccc}
\toprule
\textbf{Model} & \textbf{Avg} & \rotatebox{90}{\textbf{FiQA}} & \rotatebox{90}{\textbf{NFCorpus}} & \rotatebox{90}{\textbf{TREC-COVID}} & \rotatebox{90}{\textbf{Touche}} & \rotatebox{90}{\textbf{ArguAna}} & \rotatebox{90}{\textbf{Quora}} & \rotatebox{90}{\textbf{SCIDOCS}} & \rotatebox{90}{\textbf{SciFact}} & \rotatebox{90}{\textbf{NQ}} & \rotatebox{90}{\textbf{ClimateFEVER}} & \rotatebox{90}{\textbf{HotpotQA}} & \rotatebox{90}{\textbf{DBPedia}} & \rotatebox{90}{\textbf{CQADupstack}} & \rotatebox{90}{\textbf{FEVER}} & \rotatebox{90}{\textbf{MSMARCO}} \\
\midrule
ModernBERT-embed-unsupervised \textbf{with} prompts & 47.05 & 42.53 & 35.33 & 68.44 & 18.58 & 48.82 & 88.63 & 19.83 & 72.30 & 46.32 & 22.97 & 60.00 & 37.97 & 42.40 & 67.39 & 34.23 \\
\quad Supervised + KD \textbf{with} prompts & 55.12 & 41.50 & 36.51 & 77.46 & 33.77 & 52.45 & 86.26 & 18.66 & 74.90 & 62.24 & 37.27 & 80.07 & 48.27 & 41.60 & 89.71 & 46.17 \\
\quad Supervised + KD \textit{without} prompts & 54.44 & 41.30 & 36.00 & 79.23 & 33.52 & 52.27 & 87.52 & 18.28 & 74.90 & 61.35 & 33.25 & 77.64 & 47.40 & 41.18 & 86.57 & 46.15 \\

\midrule
ColBERT-Zero-unsupervised \textbf{with} prompts & 51.44 & 45.38 & 36.88 & 67.82 & 22.59 & 51.53 & 87.78 & 22.30 & 76.76 & 58.80 & 24.24 & 68.29 & 43.16 & 45.76 & 81.58 & 38.78 \\
\quad Supervised + KD \textbf{with} prompts & 55.43 & 42.62 & 37.28 & 78.69 & 36.13 & 53.07 & 85.24 & 19.88 & 76.50 & 61.66 & 35.72 & 79.41 & 47.48 & 41.34 & 90.59 & 45.80 \\
\quad Supervised + KD \textit{without} prompts & 54.17 & 42.99 & 37.49 & 71.95 & 27.90 & 49.29 & 88.26 & 20.29 & 76.85 & 62.20 & 32.77 & 77.91 & 47.43 & 44.67 & 89.36 & 43.14 \\

\midrule
ColBERT-Zero-unsupervised \textit{without} prompts & 51.70 & 45.31 & 34.72 & 73.55 & 23.26 & 52.56 & 88.15 & 22.63 & 76.1 & 59.18 & 24.24 & 66.66 & 42.61 & 45.56 & 81.88 & 39.15 \\
\quad Supervised + KD \textbf{with} prompts & 54.18 & 42.75 & 37.28 & 77.86 & 35.19 & 49.73 & 87.61 & 18.97 & 75.43 & 61.64 & 30.75 & 77.32 & 47.48 & 39.72 & 84.59 & 46.36 \\
\quad Supervised + KD \textit{without} prompts & 54.61 & 43.14 & 36.60 & 78.60 & 36.36 & 49.49 & 88.05 & 19.13 & 76.42 & 61.73 & 32.70 & 76.99 & 47.69 & 40.21 & 85.97 & 46.01 \\

\bottomrule
\end{tabular}%
}
\end{table*}

\section{Discussions}
\label{sec:discussions}
As started previously, we used Nomic Embed data to enable direct comparison with existing models. While this choice grounds our results in a well-established setup, it also means that some observations may not generalize to different or stronger training configurations, an avenue we leave for future work.
For example, when performing the supervised fine-tuning phase on stronger data (NV-Retriever~\cite{DBLP:journals/corr/abs-2407-15831} on various datasets), adding prompts on top of a ColBERT model pre-trained without prompts did not degrade the results as it was the case for ColBERT-Zero. With stronger/longer fine-tuning, the alignment with pre-training is less required as the model has enough room to adapt. Likewise, the small remaining advantage of pre-training a ColBERT model, rather than relying solely on fine-tuning and KD, is likely influenced by the quality and scale of those subsequent phases. Given that knowledge distillation typically \emph{improves} results after supervised fine-tuning, the fact that a prior contrastive phase further enhances performance suggests that these gains are less about the specific objective and more about the \emph{scale of training}. This extra capacity allows the model more time to optimize within a multi-vector setting. Consequently, performing KD alone at a larger scale might yield similar, if not superior, results due to the higher quality of the distillation signal. The goal of this study, however, was to explore the conventional setup where training scale is inversely proportional to signal quality, reflecting the higher cost of generating high-quality labels. Thus, the results of each phase primarily highlight the impact of \emph{scale} within those phases rather than the specific training objective itself.
\section{Conclusion}
% We compared various setups to train multi-vector models, from the usual small KD step on top of a dense-pretrained model to full pre-training. We found that the usual setup is too shallow and leaves a lot of performance on the table, but that investing in a supervised step yields performance close to full pre-training without requiring the most costly unsupervised phase. Performing these additional steps yields models that outperform models that leverages much stronger data and set new state-of-the-art.
We compared various setups for training multi-vector models, ranging from the conventional small KD step applied on top of a dense pre-trained model to full pre-training. We found that the conventional approach is lacking in depth, resulting in a significant untapped potential in terms of performance. However, we observed that investing in a supervised step leads to performance that approaches the level achieved through full pre-training, without the necessity of the most expensive unsupervised phase. This results in models that demonstrate superior performance in comparison to those leveraging stronger data and establish a new state-of-the-art.
We also investigated the impact of prompts on model performances. We found that, in the context of repurposing pre-trained models, it is imperative to meticulously align the pre-training step with respect to the use, or non-use, of prompts. We showed that the utilization of full prompts results in a positive impact in comparison with the employment of single token query and document markers. These findings present a range of interesting research avenues. To help explore multi-vector pre-training and those avenues, we publicly release \href{https://huggingface.co/collections/lightonai/colbert-zero}{all intermediate checkpoints}, including unsupervised and supervised stages with and without prompts. The full \href{https://github.com/lightonai/pylate/tree/main/examples/train/ColBERT-zero/}{training scripts}, built on PyLate and leveraging public data, are also available. We hope these resources will help the community explore the many open questions raised by our findings, from the role of prompts to the scaling behavior of multi-vector training.

\paragraph{Acknowledgments}
This work was supported as part of the Swiss AI Initiative by a grant from the Swiss National Supercomputing Centre (CSCS) under project \texttt{IDa120} on \href{https://www.cscs.ch/computers/alps}{Alps}.
We also want to thank Raphaël Sourty for the helpful discussions and reviewing the early version of this work.
%%
%% Define the bibliography file to be used
\bibliography{sample-ceur}

\appendix

\section{Discussion on the Effect of the Prompt} 
\label{appendix:prompts}
In this appendix, we provide a detailed analysis of the impact of prompts on model performance and the secondary effects of increased sequence lengths for queries and documents. We investigate whether the performance gains observed in the main text are merely a \emph{structural effect} of processing additional tokens, or if the prompts trigger more complex behaviors such as \emph{query expansion} as discussed in Section~\ref{sec:prompts}. To isolate these variables, we conducted a series of ablations by fine-tuning different models with and without the prompts (\textbf{+Prompt} setting) as well as with and without accounting for token-length parity (\textbf{+Length} setting), that is, increasing the query/document lengths by the length of the prompts, 7 tokens. All the results are reported in Table~\ref{tab:full-ablation-results} and a synthesis of the effect of each parameter for different base model on the BEIR average is presented in Table~\ref{tab:delta-heatmap}.
As discussed in the main body, matching the pre-training setup is crucial for downstream performance. Incorporating prompts during the fine-tuning of pre-trained models with prompts enhances results, while including them for models pre-trained without prompts leads to a decline. Notably, the most significant relative gains occur when performing only the KD phase, further underscoring the pivotal role of alignment. We observe that in the supervised + KD scenario, the model has more opportunity to adapt to initial misalignment, whereas the \emph{KD-only} setting is far more sensitive to the pre-training configuration. Crucially, however, alignment alone does not account for all the observed gains. As detailed in Section~\ref{sec:prompts}, comparing full ColBERT pre-training with and without prompts isolates the effect of prompts from any alignment confound, since both models are evaluated after end-to-end training in their respective settings. \textbf{This confirms that prompts provide an intrinsic benefit beyond simply matching the fine-tuning setup to the pre-training one}. 

When trying to disentangle the effect of the prompts ``content" and the increased length, we observe gains by adding the prompts even when not accounting for the additional tokens during the training of the models. While changing the length during evaluation only does not change the results (we cover the lengths of the BEIR benchmark), we conducted additional testing to determine the impact of increasing the query/document length during training. It appears that increasing the length is beneficial, even without incorporating the prompts. Again, in this case, the most significant gains are achieved by focusing solely on the KD. This phenomenon can be attributed to the fact that supplementary tokens offer extra training signals, a feature that is particularly advantageous for brief training. 

Thus, it appears that both factors explain the performance boost. These factors seem to be additive, as incorporating both increases performance beyond either individual component. Interestingly, while this holds true whether using a dense or a ColBERT model as the unsupervised base, the prompt-pre-trained ColBERT model benefits significantly more from their combination. Indeed, while the gain for the dense model is smaller than the sum of the individual contributions, the combined gain for the fully pre-trained ColBERT model is much larger than the sum of its individual gains. 

Our running theory is that the observed performance boost stems from a synergy between \emph{structural alignment} and \emph{representational capacity}. We conjecture that ColBERT, unlike dense models, treats the pre-training configuration, specifically the combination of prompts and extended sequence length, as a unified template. Because the query budget is relatively short, introducing a prompt without increasing length at the supervised stage creates a \emph{token-budget trade-off} where the prompt displacement of content tokens neutralizes the alignment benefits. The synergistic gain only emerges when the prompts provide the `instruction" and the additional length provides the ``semantic real estate" necessary for the model to execute pre-trained behaviors like \emph{implicit query expansion} without signal interference. However, this remains a preliminary conjecture; further investigation is required using a wider range of prompt variations and incremental length scales to definitively isolate the individual and combined mechanisms at play.

\newcommand{\cmark}{\ding{51}}  % checkmark
\newcommand{\xmark}{\ding{55}}  % crossmark

\begin{table*}[htbp]
\centering
\caption{Retrieval performance on the BEIR benchmark studying the impact of adding prompts and increasing query/document lengths.}
\label{tab:full-ablation-results}
\resizebox{\textwidth}{!}{%
\begin{tabular}{l cc | c | ccccccccccccccc}
\toprule
& \multicolumn{2}{c|}{\textbf{Settings}} & & \multicolumn{15}{c}{\textbf{Datasets}} \\
\cmidrule(lr){2-3} \cmidrule(lr){5-19}
\textbf{Method} & \textbf{+Prompt} & \textbf{+Length} & \textbf{Avg} & \textbf{FiQA} & \textbf{NFC} & \textbf{COV} & \textbf{Tou} & \textbf{Arg} & \textbf{Quo} & \textbf{SciD} & \textbf{SciF} & \textbf{NQ} & \textbf{Clim} & \textbf{Hot} & \textbf{DBP} & \textbf{CQA} & \textbf{FEV} & \textbf{MSM} \\
\midrule
\multicolumn{19}{l}{\textbf{\textit{KD from dense supervised (with prompts)}}} \\
\midrule
\multirow{4}{*}{Distilled} 
  & \xmark & \xmark & 52.58 & 41.58 & 36.33 & 78.69 & 33.30 & 50.16 & 82.59 & 18.01 & 75.15 & 60.30 & 25.00 & 75.00 & 45.92 & 39.39 & 82.27 & 45.06  \\
& \cmark & \xmark & 53.83 & 41.74 & 36.87 & 79.65 & 34.52 & 51.06 & 87.31 & 18.41 & 76.06 & 61.24 & 27.56 & 75.38 & 47.48 & 40.43 & 84.24 & 45.47 \\
& \xmark & \cmark & 53.28 & 41.99 & 36.41 & 78.79 & 36.76 & 47.77 & 87.51 & 18.18 & 75.21 & 60.56 & 25.63 & 76.34 & 45.83 & 39.61 & 83.42 & 45.16 \\
& \cmark & \cmark & \textbf{54.09} & 42.51 & 37.01 & 79.52 & 34.58 & 51.75 & 87.67 & 18.15 & 75.04 & 61.45 & 28.31 & 76.70 & 47.54 & 40.68 & 84.82 & 45.57 \\
\midrule
\multicolumn{19}{l}{\textit{\textbf{Supervised + KD from dense unsupervised (with prompts)}}} \\
\midrule
\multirow{4}{*}{Supervised} 
  & \xmark & \xmark & 51.33 & 42.30 & 34.52 & 73.97 & 25.22 & 49.66 & 87.67 & 17.94 & 73.57 & 60.14 & 28.53 & 72.13 & 42.54 & 42.90 & 74.04 & 44.80 \\
& \cmark & \xmark & 51.08 & 41.34 & 36.04 & 73.57 & 26.61 & 42.98 & 87.25 & 17.88 & 70.80 & 60.59 & 30.52 & 71.22 & 44.01 & 43.31 & 74.53 & 45.58 \\
& \xmark & \cmark & 51.08 & 40.93 & 35.49 & 70.56 & 22.59 & 48.96 & 87.56 & 17.50 & 71.99 & 60.50 & 32.61 & 72.90 & 43.47 & 42.10 & 73.98 & 45.07 \\
& \cmark & \cmark & 50.72 & 40.09 & 35.56 & 71.12 & 25.53 & 44.27 & 86.96 & 18.19 & 73.78 & 58.89 & 32.95 & 71.49 & 43.23 & 42.55 & 70.51 & 45.72 \\
\cmidrule(lr){1-19}
\multirow{4}{*}{Distilled} 
  & \xmark & \xmark & 54.44 & 41.30 & 36.00 & 79.23 & 33.52 & 52.27 & 87.52 & 18.28 & 74.90 & 61.35 & 33.25 & 77.64 & 47.40 & 41.18 & 86.57 & 46.15 \\
& \cmark & \xmark & 54.95 & 42.19 & 36.73 & 79.37 & 36.10 & 52.51 & 87.31 & 18.80 & 75.57 & 61.66 & 33.67 & 78.05 & 47.77 & 41.03 & 87.52 & 45.96 \\
& \xmark & \cmark & 54.82 & 42.36 & 36.33 & 80.11 & 37.47 & 48.51 & 87.74 & 18.11 & 75.92 & 61.58 & 34.46 & 78.31 & 47.27 & 40.59 & 87.21 & 46.39 \\
& \cmark & \cmark & \textbf{55.12} & 41.50 & 36.51 & 77.46 & 33.77 & 52.45 & 86.26 & 18.66 & 74.90 & 62.24 & 37.27 & 80.07 & 48.27 & 41.60 & 89.71 & 46.17 \\
\midrule
\multicolumn{19}{l}{\textit{\textbf{Supervised + KD from ColBERT unsupervised (with prompts)}}} \\
\midrule
\multirow{3}{*}{Supervised} 
  & \xmark & \xmark & 52.47 & 43.52 & 36.43 & 66.13 & 22.70 & 47.95 & 87.74 & 19.51 & 75.89 & 62.27 & 31.92 & 72.30 & 46.01 & 44.82 & 86.74 & 43.13 \\
  & \cmark & \xmark & 52.16 & 43.02 & 35.13 & 71.30 & 21.94 & 44.66 & 88.22 & 21.19 & 74.31 & 61.71 & 30.65 & 70.46 & 46.97 & 45.08 & 84.94 & 42.78 \\
& \xmark & \cmark & 52.70 & 43.59 & 36.23 & 65.03 & 23.16 & 49.89 & 87.70 & 19.61 & 75.66 & 62.34 & 33.22 & 73.46 & 45.64 & 44.92 & 87.00 & 43.12 \\
& \cmark & \cmark & 51.81 & 42.45 & 35.60 & 74.72 & 23.83 & 41.81 & 87.19 & 19.85 & 73.71 & 61.95 & 35.01 & 71.37 & 46.20 & 45.16 & 72.61 & 45.68 \\
\cmidrule(lr){1-19}
\multirow{4}{*}{Distilled} 
  & \xmark & \xmark & 54.17 & 42.99 & 37.49 & 71.95 & 27.90 & 49.29 & 88.26 & 20.29 & 76.85 & 62.20 & 32.77 & 77.91 & 47.43 & 44.67 & 89.36 & 43.14 \\
& \cmark & \xmark & 54.59 & 42.18 & 37.08 & 80.37 & 37.31 & 51.12 & 84.71 & 19.81 & 76.39 & 62.02 & 29.79 & 76.50 & 46.88 & 40.85 & 88.72 & 45.12 \\
& \xmark & \cmark & 54.47 & 42.50 & 37.85 & 77.94 & 35.65 & 51.42 & 86.99 & 18.88 & 75.73 & 61.60 & 31.37 & 77.97 & 46.81 & 40.50 & 87.08 & 44.83 \\
& \cmark & \cmark & \textbf{55.43} & 42.62 & 37.28 & 78.69 & 36.13 & 53.07 & 85.24 & 19.88 & 76.50 & 61.66 & 35.72 & 79.41 & 47.48 & 41.34 & 90.59 & 45.80 \\
\midrule
\multicolumn{19}{l}{\textit{\textbf{Supervised + KD from ColBERT unsupervised (without prompts)}}} \\
\midrule
\multirow{2}{*}{Supervised} 
  & \xmark & \xmark & 52.39 & 43.36 & 36.01 & 72.42 & 23.79 & 47.42 & 87.79 & 21.30 & 73.85 & 62.25 & 31.61 & 70.32 & 44.07 & 44.03 & 85.54 & 42.11 \\
& \cmark & \cmark & 52.62 & 44.67 & 36.20 & 72.68 & 24.47 & 48.83 & 88.00 & 21.27 & 72.65 & 61.85 & 32.84 & 69.79 & 44.98 & 44.84 & 84.95 & 41.34 \\
\cmidrule(lr){1-19}
\multirow{3}{*}{Distilled} 
  & \xmark & \xmark & 54.61 & 43.14 & 36.60 & 78.60 & 36.36 & 49.49 & 88.05 & 19.13 & 76.42 & 61.73 & 32.70 & 76.99 & 47.69 & 40.21 & 85.97 & 46.01 \\
& \xmark & \cmark$^\dagger$ & 54.60 & 43.25 & 36.88 & 78.67 & 36.27 & 49.14 & 88.12 & 19.06 & 76.28 & 61.58 & 32.25 & 77.55 & 47.23 & 40.34 & 86.34 & 46.06 \\
& \cmark & \cmark & 54.18 & 42.75 & 37.28 & 77.86 & 35.19 & 49.73 & 87.61 & 18.97 & 75.43 & 61.64 & 30.75 & 77.32 & 47.48 & 39.72 & 84.59 & 46.36 \\
\bottomrule
\multicolumn{19}{l}{\footnotesize $^\dagger$ Extra length applied only during evaluation.} \\
\end{tabular}%
}
\end{table*}

\begin{table}[htbp]
\centering
\caption{Average BEIR nDCG@10 with delta ($\Delta$) for adding prompts and increasing lengths vs. baseline (no prompt, no length)}
\label{tab:delta-heatmap}

\definecolor{pos3}{RGB}{80,200,80}   % Strong positive
\definecolor{pos2}{RGB}{144,238,144} % Medium positive
\definecolor{pos1}{RGB}{220,255,220} % Light positive
\definecolor{neg1}{RGB}{255,220,220} % Light negative
\definecolor{neg2}{RGB}{255,160,160} % Medium negative
\definecolor{baseline}{RGB}{240,240,240} % Gray for baseline

\newcolumntype{C}{>{\centering\arraybackslash}m{2.4cm}}

\begin{tabular}{l|C|CCC}
\toprule
\textbf{Method / Training} & \textbf{Baseline} & \textbf{+Prompt} & \textbf{+Length} & \textbf{+Both} \\
\midrule
\multicolumn{5}{l}{\textit{KD from dense supervised (with prompts)}} \\
\midrule
Distilled & \cellcolor{baseline}52.58 & \cellcolor{pos3}+1.25 (53.83) & \cellcolor{pos2}+0.70 (53.28) & \cellcolor{pos3}\textbf{+1.51 (54.09)} \\
\midrule
\multicolumn{5}{l}{\textit{Supervised + KD from dense unsupervised (with prompts)}} \\
\midrule
Supervised & \cellcolor{baseline}\textbf{51.33} & \cellcolor{neg1}--0.25 (51.08) & \cellcolor{neg1}--0.25 (51.08) & \cellcolor{neg2}--0.61 (50.72) \\
Distilled & \cellcolor{baseline}54.44 & \cellcolor{pos1}+0.51 (54.95) & \cellcolor{pos1}+0.38 (54.82) & \cellcolor{pos2}\textbf{+0.68 (55.12)} \\
\midrule
\multicolumn{5}{l}{\textit{Supervised + KD from ColBERT unsupervised (with prompts)}} \\
\midrule
Supervised & \cellcolor{baseline}52.47 & \cellcolor{neg1}--0.31 (52.16) & \cellcolor{pos1}\textbf{+0.23 (52.70)} & \cellcolor{neg2}--0.66 (51.81) \\
Distilled & \cellcolor{baseline}54.17 & \cellcolor{pos1}+0.42 (54.59) & \cellcolor{pos1}+0.30 (54.47) & \cellcolor{pos3}\textbf{+1.26 (55.43)} \\
\midrule
\multicolumn{5}{l}{\textit{Supervised + KD from ColBERT unsupervised (without prompts)}} \\
\midrule
Supervised & \cellcolor{baseline}52.39 & \multicolumn{2}{c}{--} & \cellcolor{pos1}\textbf{+0.23 (52.62)} \\
Distilled & \cellcolor{baseline}\textbf{54.61} & \multicolumn{2}{c}{--} & \cellcolor{neg1}--0.43 (54.18) \\
\bottomrule
\end{tabular}

% t\vspace{1em}
% \footnotesize
% \textcolor{pos3}{$\blacksquare$} $\Delta > 1.0$ \quad
% \textcolor{pos2}{$\blacksquare$} $0.5 < \Delta \leq 1.0$ \quad
% \textcolor{pos1}{$\blacksquare$} $0 < \Delta \leq 0.5$ \quad
% \textcolor{neg1}{$\blacksquare$} $-0.5 \leq \Delta < 0$ \quad
% \textcolor{neg2}{$\blacksquare$} $\Delta < -0.5$

\end{table}

\end{document}